\documentclass[pmlr,twocolumn, breaklinks]{jmlr}

\usepackage{longtable}

\usepackage{booktabs}

\usepackage[load-configurations=version-1]{siunitx} 
\usepackage{cuted}
\usepackage{soul}

\theorembodyfont{\upshape}
\theoremheaderfont{\scshape}
\theorempostheader{:}
\theoremsep{\newline}

\usepackage[nolist]{acronym}

\makeatletter
\newcommand*{\org@overidelabel}{}
\let\org@overridelabel\@verridelabel
\@ifpackagelater{acronym}{2015/03/21}{
  \renewcommand*{\@verridelabel}[1]{%
    \@bsphack
    \protected@write\@auxout{}{\string\AC@undonewlabel{#1@cref}}%
    \org@overridelabel{#1}%
    \@esphack
  }%
}{
  \renewcommand*{\@verridelabel}[1]{%
    \@bsphack
    \protected@write\@auxout{}{\string\undonewlabel{#1@cref}}%
    \org@overridelabel{#1}%
    \@esphack
  }%
}
\makeatother

\usepackage{url}

\usepackage{hyperref}

\title[Recent Advances, Applications and Open Challenges from ML4H Roundtables]{Recent Advances, Applications and Open Challenges in Machine Learning for Health: Reflections from  Research Roundtables at ML4H 2023 Symposium}

 \author{
   \Name{Hyewon Jeong\nametag{\textsuperscript{1}}} \Email{hyewonj@mit.edu}\vspace{-0.095cm}\\
   \Name{Sarah Jabbour\nametag{\textsuperscript{1, 2}}} \Email{sjabbour@umich.edu}\vspace{-0.095cm}\\
   \Name{Yuzhe Yang\nametag{\textsuperscript{1}}} \Email{yuzhe@mit.edu}\vspace{-0.095cm}\\
   \Name{Rahul Thapta\nametag{\textsuperscript{2}}} \Email{rthapa84@stanford.edu}\vspace{-0.095cm}\\
   \Name{Hussein Mozannar\nametag{\textsuperscript{2}}} \Email{mozannar@mit.edu}\vspace{-0.095cm}\\
   \Name{William Jongwon Han\nametag{\textsuperscript{2}}} \Email{wjhan@andrew.cmu.edu}\vspace{-0.095cm}\\
    \Name{Nikita Mehandru\nametag{\textsuperscript{2}}} \Email{nmehandru@berkeley.edu}\vspace{-0.095cm}\\
    \Name{Michael Wornow\nametag{\textsuperscript{2}}} \Email{mwornow@stanford.edu}\vspace{-0.095cm}\\
   \Name{Vladislav Lialin\nametag{\textsuperscript{2}}} \Email{vlad.lialin@gmail.com}\vspace{-0.095cm}\\
   \Name{Xin Liu\nametag{\textsuperscript{2}}} \Email{xliucs@google.com}\vspace{-0.095cm}\\
   \Name{Alejandro Lozano\nametag{\textsuperscript{2}}} \Email{lozanoe@stanford.edu}\vspace{-0.095cm}\\
   \Name{Jiacheng Zhu\nametag{\textsuperscript{2}}} \Email{zjc@mit.edu}\vspace{-0.095cm}\\
   \Name{Rafal Dariusz Kocielnik\nametag{\textsuperscript{2}}} \Email{rafalko@caltech.edu}\vspace{-0.095cm}\\
  \Name{Keith Harrigian\nametag{\textsuperscript{2}}} \Email{kharrigian@jhu.edu}\vspace{-0.095cm}\\
  \Name{Haoran Zhang\nametag{\textsuperscript{2}}} \Email{haoranz@mit.edu}\vspace{-0.095cm}\\
  \Name{Edward Lee\nametag{\textsuperscript{2}}} \Email{edhlee@stanford.edu}\vspace{-0.095cm}\\
  \Name{Milos Vukadinovic\nametag{\textsuperscript{2}}} \Email{milosvuk@g.ucla.edu}\vspace{-0.095cm}\\
  \Name{Aparna Balagopalan\nametag{\textsuperscript{2}}} \Email{aparnab@mit.edu}\vspace{-0.095cm}\\
  \Name{Vincent Jeanselme\nametag{\textsuperscript{2}}} \Email{vincent.jeanselme@mrc-bsu.cam.ac.uk}\vspace{-0.095cm}\\
  \Name{Katherine Matton\nametag{\textsuperscript{2}}} \Email{kmatton@mit.edu}\vspace{-0.095cm}\\
  \Name{Ilker Demirel\nametag{\textsuperscript{2}}} \Email{demirel@mit.edu}\vspace{-0.095cm}\\
   \Name{Jason Fries\nametag{\textsuperscript{3}}} \Email{jfries@stanford.edu}\vspace{-0.095cm}\\
   \Name{Parisa Rashidi\nametag{\textsuperscript{3}}} \Email{parisa.rashidi@bme.ufl.edu}\vspace{-0.095cm}\\
   \Name{Brett Beaulieu-Jones\nametag{\textsuperscript{3}}} \Email{beaulieujones@uchicago.edu}\vspace{-0.095cm}\\
   \Name{Xuhai Orson Xu\nametag{\textsuperscript{3}}} \Email{xoxu@mit.edu}\vspace{-0.095cm}\\
   \Name{Matthew McDermott\nametag{\textsuperscript{3}}} \Email{matthew\_mcdermott@hms.harvard.edu}\vspace{-0.095cm}\\
   \Name{Tristan Naumann\nametag{\textsuperscript{3}}} \Email{tristan@microsoft.com}\vspace{-0.095cm}\\
   \Name{Monica Agrawal \nametag{\textsuperscript{3}}} \Email{magrawal@mit.edu}\vspace{-0.095cm}\\
   \Name{Marinka Zitnik\nametag{\textsuperscript{3}}} \Email{marinka@mit.edu}\vspace{-0.095cm}\\
   \Name{Berk Ustun\nametag{\textsuperscript{3}}} \Email{berk@ucsd.edu}\vspace{-0.095cm}\\
   \Name{Edward Choi\nametag{\textsuperscript{3}}} \Email{edwardchoi@kaist.ac.kr}\vspace{-0.095cm}\\
   \Name{Kristen Yeom\nametag{\textsuperscript{3}}} \Email{kyeom@stanford.edu}\vspace{-0.095cm}\\
   \Name{Gamze Gürsoy\nametag{\textsuperscript{3}}} \Email{gg2845@cumc.columbia.edu}\vspace{-0.095cm}\\
   \Name{Marzyeh Ghassemi\nametag{\textsuperscript{3}}} \Email{mghassem@mit.edu}\vspace{-0.095cm}\\
   \Name{Emma Pierson\nametag{\textsuperscript{3}}} \Email{ep432@cornell.edu}\vspace{-0.095cm}\\
   \Name{George Chen\nametag{\textsuperscript{3}}} \Email{georgechen@cmu.edu}\vspace{-0.095cm}\\
   \Name{Sanjat Kanjilal\nametag{\textsuperscript{3}}} \Email{skanjilal@bwh.harvard.edu}\vspace{-0.095cm}\\
   \Name{Michael Oberst\nametag{\textsuperscript{3}}} \Email{moberst@andrew.cmu.edu}\vspace{-0.095cm}\\
   \Name{Linying Zhang\nametag{\textsuperscript{3}}} \Email{linyingz@wustl.edu}\vspace{-0.095cm}\\
    \Name{Harvineet Singh\nametag{\textsuperscript{1}}} \Email{hs3673@nyu.edu}\vspace{-0.095cm}\\
   \Name{Tom Hartvigsen\nametag{\textsuperscript{1}}} \Email{tomh@mit.edu}\vspace{-0.095cm}\\
   \Name{Helen Zhou\nametag{\textsuperscript{1}}} \Email{hlzhou@andrew.cmu.edu}\vspace{-0.095cm}\\
   \Name{Chinasa T. Okolo\nametag{\textsuperscript{1}}} \Email{cto9@cornell.edu}\\
 \addr\textsuperscript{1}Organizing committee for ML4H 2023 Research Roundtables, \textsuperscript{2}Junior Chairs, \textsuperscript{3}Senior Chairs \\
 \vspace{-1.5cm}
   }

\begin{document}
\begin{acronym}
    \acro{ML4H}{Machine Learning for Health}
    \acro{DL}{deep learning}
    \acro{EHR}{electronic health record}
    \acro{LLM}{large language model}
    \acro{OOD}{out-of-distribution}
    \acro{RCT}{Randomized Clinical Trials}
\end{acronym}

\maketitle
\clearpage      

%
%
%

\section{Introduction}
\label{sec:intro}

The third \ac{ML4H} symposium was held in person on December 10, 2023, in New Orleans, Louisiana, USA~\citep{pmlr-v193-parziale22a}. The symposium included research roundtable sessions to foster discussions between participants and senior researchers on timely and relevant topics for the \ac{ML4H} community. Encouraged by the successful virtual roundtables in the previous year~\citep{roy2021machine}, we organized eleven in-person roundtables and four virtual roundtables at \ac{ML4H} 2022~\citep{pmlr-v193-parziale22a, hegselmannrecent}. The organization of the research roundtables at the conference involved 17 Senior Chairs and 19 Junior Chairs across 11 tables. Each roundtable session included invited senior chairs (with substantial experience in the field), junior chairs (responsible for facilitating the discussion), and attendees from diverse backgrounds with interest in the session's topic. Herein we detail the organization process and compile takeaways from these roundtable discussions, including recent advances, applications, and open challenges for each topic. We conclude with a summary and lessons learned across all roundtables. This document serves as a comprehensive review paper, summarizing the recent advancements in machine learning for healthcare as contributed by foremost researchers in the field.

\section{Organization Process}
\label{sec:organizationprocess}
An initial set of topics were identified from papers in the broad ML for health literature published in the last three years suggestions from \ac{ML4H} chairs and keynote speakers. After removing duplicates, there were $18$ unique topic candidates. For each of the topics, we invited senior chairs with expertise in the respective area, aiming for two chairs for each roundtable. Next, we identified junior chairs that preferably had some experience in the discussed topic. Before the event, junior and senior chairs wrote an introduction paragraph shared on the \ac{ML4H} website\footnote{https://ml4health.github.io/2023/} and submitted three to five potential discussion questions. On the day of the symposium, we had two 25-minute slots for roundtables with a five-minute break to allow participants to join another roundtable session. After the event, we asked the chairs to write a summary of the main takeaways from the discussion.

\section{Research Roundtables}
\label{sec:roundtables}

We included the following topics for the research roundtables of ML4H 2023 :
\begin{enumerate}
    \item Health AI Collaborations, Deployment, and Regulation
    \item Integrating AI into Clinical Workflows
    \item Health AI Foundational Models
    \item Large Language Models and Healthcare
    \item Multimodal AI for Health
    \item Health AI Model Development and Generalizability
    \item Health AI and Accessibility
    \item Health AI and Patient Privacy
    \item Bias/Fairness in Health AI
    \item ML for Survival Analysis \& Epidemiology/Population Health
    \item Causality
\end{enumerate}

In the following sections, we provide slightly revised versions of the introductions and summaries from all roundtables.

\subsection{Health AI Collaborations, Deployment, and Regulation}
\paragraph{Subtopic: } One of the barriers to deploying AI models in healthcare is the ability to safely and effectively integrate models into clinical workflows. What are the different factors one should consider in presenting AI models to clinicians that result in effective clinician-AI collaborations, and how do we know if these models truly have a significant impact within the healthcare setting? What are the desires of caregivers and clinicians, and what aspects are still lacking?  Furthermore, there has been an increased focus on AI regulation by policymakers and industry players in the last few years. How do we ensure all stakeholders are considered in AI policy, and who should be in charge of writing such regulation, if at all?

\paragraph{Chairs:}
\textit{Jason Fries, Parisa Rashidi, Hussein Mozannar, Rahul Thapta}

\paragraph{Background:}

This research round table focused on three topics. 
\begin{enumerate}

\item Health AI Collaborations: Exploring synergies and challenges in partnering between AI researchers and healthcare institutions, focusing on optimizing the interface between human expertise and AI technology to enhance healthcare outcomes.

\item Deployment: The practical aspects of implementing AI in healthcare settings, discussing the necessary infrastructure, clinician interaction with AI systems, and strategies for integration into healthcare practices.

\item Regulation: The general need for effective regulatory frameworks for AI in healthcare, covering aspects such as policy evolution \citep{mello2023president}, liability concerns \citep{mello2023chatgpt}, and balancing innovation with regulation to ensure safe and responsible use of AI in medical contexts.
\end{enumerate}

\paragraph{Discussion:}

Participants discussed several key challenges when conducting machine learning for healthcare deployment:

\paragraph{Interfacing with Clinical Experts}

Effective integration requires forming productive collaborations with clinical professionals and developing a common language to facilitate communication. An example cited was the University of Florida's development of a real-time system for data acquisition and integration into AI engines \cite{shickel2019deepsofa, ren2022performance}. Essential to this process is collaboration with IT and hospital administrators, aligning the incentives between hospital staff and computer science professionals, and considering the needs of caregivers and patients. The initial step for any research in this area is to identify a domain expert. Understanding the problem comprehensively, grasping the efficacy of the workflow, and observing clinical workflow firsthand is vital. 
Another key factor is identifying an individual within the system who can champion the initiative.

\paragraph{Data Acquisition Challenges}

The focus in machine learning often leans towards model building, while data acquisition and infrastructure development are somewhat sidelined. 
Challenges include the reluctance to add extra steps in existing workflows, the costs associated with hosting models, and the reluctance to bear these costs. Clinicians, who are already pressed for time, should not be burdened with additional data collection responsibilities.
This aligns with the well-commented data challenges in deployed machine learning systems and so-called "data cascades" that can result from undervaluing data curation and knowledge of generating processes \citep{sambasivan2021everyone}. 

\paragraph{Model Evaluation: \ac{RCT} and Local Validation}

Model evaluation processes, such as recurring internal and prospective validation paradigms \citep{youssef2023external}, and external validation (including collaboration with other institutions), each have distinct advantages and are relevant at different development stages. A crucial question is the role of external validation in an era where foundation models are shared and adapted for use at individual centers \citep{guo2023multi}.

\paragraph{Human-AI Collaboration in Evaluating Deployed Systems}

Model evaluation in the generative AI era faces many challenges.
The current paradigm of evaluating AI models in isolation is fundamentally flawed, as future models will likely function as co-pilots in clinical settings.
This requires evaluating the behavior of healthcare workers and AI systems collaborating to make clinical decisions. 
Popular current healthcare benchmarks for \ac{LLM} are moreover misaligned with how clinicians interact with AI systems, for example focusing on narrow, USMLE-style multiple choice evaluations at the cost of more complex information synthesis \citep{fleming2023medalign}. 
The focus should be on in-situ evaluation and usage effectiveness. The current vision of autonomous AI agents might be overly ambitious, emphasizing the need for understanding human-AI interaction and building trust in AI models. The user interface of AI recommendations is crucial; poor interfaces can hinder adoption. 
The evaluation of AI in healthcare should not only focus on short-term outcomes but also consider long-term patient outcomes and the potential increase in clinicians' workload.

\paragraph{Transitioning from Research Demos to Deployed Models}

There is a misunderstanding about how data is generated and collected in clinical settings versus research environments, e.g., the data feed from an Epic EHR system looks quite different than the data available to researchers via an academic data warehouse. 
Clinical AI requires the development of machine learning operations tools and a better interface for data integration in hospitals. 
Continuous monitoring is essential for deployment, and the long-term measurement of outcomes is crucial, which might necessitate the use of causal inference tools. 
All of these require advancements in deployment science and core methods development.
\subsection{Integrating AI into Clinical Workflows}
\paragraph{Subtopic:} The rate of AI progress in the last few years seems to have major implications for the types of models we train for healthcare purposes. With the types of models we train constantly changing, how can we develop model-agnostic methods to integrate AI into clinical workflows? 

\paragraph{Chairs:}
\textit{Brett Beaulieu-Jones, William Jongwon Han, Nikita Mehandru}

\paragraph{Background:} 
The rate of AI deployment in clinical settings is driven by the models we train, and its reliability has far-reaching consequences. Data variability from different modalities in diverse medical settings (e.g., safety-net hospitals, private practice, and academic institutions) makes developing model-agnostic methods to integrate AI into clinical workflows a challenging endeavor. Distribution shifts are common in medicine \citep{zech2018variable, beede2020human} due to the difficulty in obtaining heterogeneous patient data at multiple points in time as well as the prevalence of missingness \citep{ayilara2019impact, getzen2023mining}.

As a result, the lack of model generalizability can contribute to a reduction in clinician trust in AI, and signals the need for post-deployment monitoring of systems. 

\paragraph{Discussion:}
Addressing potential distribution shifts in healthcare data is difficult, but can be prevented by collecting additional data and conducting a prospective analysis as well as an assessment of clinical outcomes. One participant noted that acquiring more patient data does not solve the \ac{OOD} issue if a new medication is introduced, and randomization at the patient level may not capture sub-group differences. More robust audits prior to deployment were suggested as a technique for mitigating the failure of systems in practice. 

Furthermore, the development of uncertainty quantification measures can aid clinicians in calibrating the reliability of model performance in medical settings. We discussed the immense importance of interpretable estimates. The value of conducting need-finding interviews with clinicians before building a model was also proposed as a method for both ensuring model alignment and increasing adoption of AI systems, as greater confidence is crucial to mediate the liability clinicians carry if systems fail. Lastly, these conversations cannot occur in a vacuum, and linking clinical outcomes to business incentives was stressed, as hospitals are often evaluated on such objectives. 

\subsection{Health AI Foundation Models}
\paragraph{Subtopic:} Foundation models train on large amounts of data, which might require sourcing data from multiple hospitals. How should we think about the data that drives the development and deployment of such models in health systems?

\paragraph{Chairs:}
\textit{Matthew McDermott, Tristan Naumann, Michael Wornow, Vlad Lialin}

\paragraph{Background:} Hospitals generate an average of 50 petabytes of data per year, over double the size of the US Library of Congress's collections \citep{weforum}. Unfortunately, the vast majority of this data remains unused --- there is simply too much data and too few resources to clean, label, and build upon it \citep{weforum, healthcarefinancenewsMostData}. Foundation models -- large-scale machine learning models trained on vast amounts of unlabeled data -- offer a promising approach for harnessing this untapped wealth of data \citep{bommasani2021opportunities}.

Unfortunately, significant practical challenges remain in building foundation models for healthcare data. First, the biggest recent successes in AI have revolved been \textbf{\acp{LLM}}, which require Internet-scale data to be successfully trained \citep{gao2020pile}. Pooling healthcare data across hospitals remains difficult due to a lack of shared data standards, misaligned incentives, and patient privacy concerns \citep{mattioli2017data}. Second, in addition to the challenges of sourcing the appropriate scale of data, questions remain about which data modalities to include and how to include them (e.g., clinical notes, structured \ac{EHR} data, images, genetic tests, etc.) \citep{moor2023foundation}. Third, it is unclear what capabilities we might expect (or want) out of such a foundation model were we to successfully train one of sufficient scale where we might observe emergent properties \citep{wei2022emergent}.

\paragraph{Discussion:} Among many other topics, our conversation surfaced several potential answers to the three core challenges listed above.

On the topic of data standards, there has been a recent push across academic research labs towards developing shared frameworks for processing health data into a machine-learning-friendly format \citep{mcdermott2023esgpt, wang2020mimic-extract, karargyris2023federated}. During our discussion, many participants advocated for the development of a common set of infrastructure upon which the ML for the healthcare community can confidently build models of the architecture and scale necessary to unlock the full potential of foundation models. However, questions were raised as to whether the key obstacle here was technical, or whether first clearing the political, regulatory, and financial roadblocks is a necessary prerequisite. We did note that even though many of the most powerful clinical foundation models published \citep{Med-PaLM-2, Med-PaLM-M, Med-Flamingo, li2023llavamed} were not trained on clinical data and instead relied on publicly available biomedical texts \citep{wornow2023shaky}, the ability of these models to rapidly learn from high-quality data opens up avenues for more data-efficient model training that sidesteps many of these data aggregation issues \cite{gunasekar2023textbooks}. 

In addition to the question of aggregating sufficient data, much of our discussion was spent on the types of data that should be fed into such a model. Participants noted that multimodal health data encompasses a much broader spectrum of inputs than what general ``multimodal'' models are expected to ingest (i.e., text, audio, and images) \cite{team2023gemini}. Healthcare data could include not only text, images, audio, and video, but also -omics data (e.g., genomics, metabolomics, transcriptomics, etc.), structured EHR data (e.g., billing, drug, and procedure codes), and waveform vitals. The complexity of the types of multimodal reasoning performed by a physician also makes the development of multimodal healthcare models complex. However, we were encouraged by the recent development of multimodal models for clinical decision-making support \citep{Med-PaLM-2, Med-PaLM-M, Med-Flamingo,li2023llavamed,zhang2023biomedclip} and molecular biology \citep{nguyen2023hyenadna, dalla2023nucleotide,lin2023evolutionary}. 

Finally, we discussed the topic of ``emergence'' and asked what sort of novel capabilities might emerge from a healthcare foundation model at a sufficient scale. While we noted that the concept of ``emergence'' is still under debate \citep{schaeffer2023emergent}, we were excited by the nascent capabilities demonstrated by large-scale models fine-tuned for clinical applications \citep{Med-PaLM-2, Med-PaLM-M}, and concluded that our ability to imagine the ways in which these technologies might one day be applied in the clinic is likely highly myopic. The emergence of unprecedented capabilities also poses risks, and the holistic evaluation of such foundation models when deployed in healthcare settings will become increasingly important to mitigate risk to patients and providers \citep{wornow2023ehrshot, liang2023helm}.

\subsection{Large Language Models and Healthcare}

\paragraph{Subtopic:} What are some low-hanging fruit opportunities to use large language models in healthcare?

\paragraph{Chairs:}
\textit{Monica Agarwal, Xin Liu, Alejandro Lozano}

\paragraph{Background:} 

In recent years, the intersection of \acp{LLM}, consumer health applications,  and the broader field of medicine has emerged as a compelling area of research with profound implications. \acp{LLM}, exemplified by advanced natural language processing technologies, have demonstrated their capacity to revolutionize various industries, including healthcare. The convergence of these technologies holds the promise of unlocking new frontiers in data-driven decision-making, personalized health tracking, and improved communication within the healthcare ecosystem.


\paragraph{Discussion:} 

Our table discussion started with recent advances in the field. We emphasized the current limitations of the field, with a special focus on meaningful evaluations and the possible risks of such models. Regarding evaluations, we agreed we need to move on from multiple-choice exam-style questions and truly asses \acp{LLM}. This echoes the idea of "Will advancements in the field improve clinical care?" A special concern was the misuse of such models for misinformation and the risks such models could pose for self-medication.

\subsection{Multimodal AI for Health}

\paragraph{Subtopic:} How do we effectively integrate multiple data sources (e.g., Electronic Health Records (EHRs), images, genomics) for ML applications in healthcare? How does this work in real-time in a hospital?

\paragraph{Chairs:}
\textit{Marinka Zitnik, Jiacheng Zhu, Rafal Dariusz Kocielnik}

\paragraph{Background:} 
Multimodal AI in healthcare is an emerging field that harnesses the synergies of diverse data sources, such as \acp{EHR}, images, and genomics \citep{kline2022multimodal, ektefaie2023multimodal}. This approach moves beyond the confines of single data modalities \citep{mcdermott2020comprehensive, jeong2023deep, jeong2024event}, aiming to enhance diagnostic accuracy and treatment efficacy. However, this integration brings forth challenges in technological integration, data management, and ethical handling, particularly in protecting the privacy of sensitive health data as well as challenges of unequal data availability for certain demographics and geographical areas.

Recent studies highlight the potential and challenges of multimodal AI in healthcare. Showing that the use of multimodal methods in neurology and oncology can increase predictive accuracy \citep{kline2022multimodal}. Further work in oncology has shown that integrating various modalities can contribute to biomarker discovery and reaching therapeutic targets \citep{lipkova2022artificial}. 
Beyond oncology, multimodal approaches have been shown particularly useful for disease relation extraction \citep{lin2023multimodal}, surgical feedback classification \citep{kocielnik2023deep}, medical record retrieval for electrocardiogram (ECG) \citep{qiu2023automated}, risk stratification with continuous monitors \citep{raghu2023sequential}, as well as pulmonary embolism detection through integration of CT imaging and EMR data \citep{huang2020multimodal}.

On the technical side, techniques for combining multiple modalities involved early, intermediate, and late fusion approaches \citep{boulahia2021early} as well as graph-based architectures \citep{lin2023multimodal} among others.

At the same time \cite{ektefaie2023multimodal}  discuss the challenges posed by heterogeneous graph datasets in multimodal graph AI methods and \cite{kline2022multimodal} highlight the challenges related to the lack of consensus on the optimal way of combining multimodal data, the sensitivity of the best architecture to disease/application, as well as increased challenges in the interpretability of model development.

These studies collectively indicate the diverse applications and progress in Multimodal AI for Health, while also highlighting the complexities involved in integrating different data types. As the field continues to evolve, the integration of various data sources in healthcare presents significant opportunities for advancement. However, it also poses notable challenges, particularly in the practical deployment of these systems in real-time clinical environments.

\paragraph{Discussion:} In the discussion panel, we focused on the key questions in this area: 1) Is more always better, or can we think of situations, patient populations, or healthcare tasks where using multiple data modalities can be counterproductive, unhelpful, or even harmful? 2) What are the algorithmic limitations of prevailing multimodal architectures particularly in Health? 3) Can multimodal learning create fundamental new capabilities in Health? 4) Are there important problems that unimodal architectures cannot solve?

In this context, the roundtable discussion focused on multimodal machine learning, its applications, and challenges, especially in the fields of healthcare and biology. The panel comprised experts from various disciplines, providing a comprehensive overview of current practices, ethical considerations, and future directions in multimodal learning. Several themes emerged:

\paragraph{Ethical Standards and Data Collection Strategies:}
This theme focused on ethical standards in multimodal learning~\citep{alwahaby2022evidence_ethical_mm}. The discussion emphasized the importance of understanding and evaluating existing frameworks, especially in the context of data collection strategies. The distinction between pre-training and fine-tuning in multi-task learning was highlighted, with a special mention of the successes in pre-training scenarios such as the development of various foundation models. 
These scenarios showcased the powerful potential of matched multimodal data for the same unit of analysis (e.g., text and vision)\citep{zhang2023biomedclip}. 
The concept of augmenting and leveraging unmatched data was also discussed, with an emphasis on learning from both text and audio data at scale. Furthermore, the discussion touched on representation learning, particularly in genomic sequencing, and the challenges posed by the increasing prevalence of genomic data. 
An important point was also raised around the availability and quality of multimodal data for various demographics and geographical regions. This was a particularly novel and important discussion point as increased reliance on high-quality multimodal sources can disenfranchise vast global populations from access to high-quality AI for Health solutions.

\textbf{Challenges in Multimodal Learning and Genomics:}
An important theme also covered the challenges of multimodal learning in bioinformatics, especially concerning omics data\citep{stahlschmidt2022multimodal_bio_reivew}. There was a discussion on the importance of incorporating prior knowledge into data handling, and the variance in genomic data based on region. One of the significant challenges noted was the development of new encoders for such data that are visually interpretable and the inherent difficulty in interpreting genomics data. 
The alignment of multimodal datasets \citep{tsai2019multimodal}, like clinical and behavioral data with genomic information, was discussed. The panel also discussed the potential of aligning gene counts and transforming them into tabular data, aiming to develop a model that can be generalized across various modalities.

\textbf{Future Directions and Dataset Development:}

The final theme focused on future directions, particularly the need for reasoning in multimodal learning and the potential for minimizing healthcare costs through machine learning. 
Discussions included the evaluation of gains from different modalities, the use of cheaper modalities as substitutes, and the challenges in inferring information across different modalities. The concept of multi-view systems, offering complementary information, was also touched upon. This discussion concluded with thoughts on the development of new evaluation benchmarks, the balance between accuracy, data quantity, and quality, as well as the potential for multi-modal language models to link knowledge across different domains. 
The value of non-traditional data resources and the creation of synthetic generative data were also noted as areas ripe for exploration.

Overall, the meeting underscored the complexities and potential of multimodal learning, especially in the context of healthcare and genomics, and highlighted the need for ethical standards, robust data collection strategies, and innovative approaches to dataset development and utilization.
\subsection{Health AI Model Development and Generalizability}

\paragraph{Subtopic:} Applying ML models in real practice could face multiple challenges including domain shift, annotation quality, and out-of-distribution. How can we ensure the robustness and generalizability of a model?

\paragraph{Chairs:}
\textit{Berk Ustun, Haoran Zhang, Keith Harrigian}

\paragraph{Background:}
Clinical machine learning models constantly face distribution shifts when deployed, which can deteriorate their ability to make accurate and calibrated predictions \citep{sendak2020path,subbaswamy2020development}. First, models may experience domain shifts when trained on a publicly available data source and deployed in a region with different population characteristics \citep{wong2021external}. Second, models continually face temporal shifts when deployed. Such shifts may occur slowly, as clinical guidelines and population characteristics change over time \citep{guo2022evaluation}. They may also be sudden, which may happen as hospital policies change \citep{nestor2019feature}, or with the introduction of a new disease such as the COVID-19 pandemic \citep{duckworth2021using}.

To effectively detect and adapt to distribution shifts, we must address several key methodological questions: (1) How can we monitor a deployed model to see if performance has dropped significantly? (2) Once a distribution shift occurs, how can we characterize \textit{how} the distribution has changed, in a way that gives insights to model building? (3) How can we build models that are resilient to distribution shifts, or update our model to accommodate these shifts? What are the assumptions or domain knowledge required?

Although distribution shift is well-studied in the core machine learning literature \citep{koh2021wilds, gulrajani2020search}, healthcare data presents many unique challenges. For example, distribution shifts in healthcare are far more difficult to describe without domain knowledge compared to natural images and text~\citep{gichoya2022ai}; bayes error may be significant and hard to measure (e.g. due to data quality issues) \citep{chen2018my} and models often contribute to high stakes decisions with a limited number of annotated training samples. 

In this roundtable, we sought to gather diverse perspectives regarding the challenges of training and evaluating robust ML models in the healthcare domain, as well as practical techniques for addressing these challenges given the constraints and nuances of healthcare data.

\paragraph{Discussion:}

We started the roundtable by asking participants about their experience with distribution shifts in real-world settings. One participant commented on the performance drops they experienced when deploying a model trained on tabular data from the intensive care unit of one hospital for risk prediction in another hospital, and using forward feature selection to identify stable and robust features \citep{heinze2018invariant}. Another participant commented on their experience with adjusting for selection bias \citep{bishop2018using} when building predictive algorithms using wearable data to obtain a model that works well for the general US population, not just those who currently own a wearable.

We next asked participants which generalization problem setup they find most useful in the healthcare setting. Compared to problem settings such as domain adaptation \citep{ben2006analysis} and domain generalization \citep{arjovsky2019invariant}, participants found single domain generalization \citep{wang2021learning} to be most useful -- a challenging scenario where only a single source domain is available, and target domain data is unobserved. 

When asked about potential solutions to the generalization problem, several participants initially raised the idea of foundation models as the silver bullet -- how pretraining on large, diverse datasets from across the world would mean that nothing is out-of-distribution during deployment. However, we identified several issues with this argument. First, while clinical foundation models have risen in popularity in text and vision \citep{tu2023towards}, they have not been widely adopted for clinical time series or tabular data. Second, the constant advent of new drugs and conditions (e.g. COVID-19) still presents a covariate shift for these models, and updating the knowledge within foundation models efficiently can be challenging \citep{zhang2024comprehensive}. Finally, a fine-tuned foundation model would still be susceptible to concept drift (e.g. due to changing clinical guidelines). 

Many participants also emphasized the importance of causality in building robust and generalizable models. Alas, the efficacy of causal reasoning as a mechanism for encouraging generalization critically depends on the validity of assumptions regarding the data generating procedure (DGP) \citep{rosenfeld2020risks} or causal graph. Participants of the roundtable consistently highlighted the importance of collaborating closely with domain experts (e.g., healthcare providers, patients, and non-ML researchers) to support these efforts. They cited personal experiences asking domain experts to propose likely sources of distribution shift in the DGP, and also noted instances in which they leveraged expert knowledge to make decisions regarding model selection (e.g., feature reduction, architecture choice). Domain experts were generally perceived and treated as a source of feedback for decisions driven predominantly by ML practitioners. Although such a relationship is unsurprising given the capacity constraints faced by many experts in the healthcare domain \citep{liebhaber2009hospital,spasic2020clinical}, emerging evidence regarding the value of human-centered design for building robust ML systems suggests it may be a dynamic worth re-evaluating \citep{auernhammer2020human,holeman2020human}.

The remaining discussion focused predominantly on technical solutions for promoting ML model generalization and ensuring reliability upon deployment. Participants highlighted two core challenges --- 1) accurately characterizing distribution shift, and 2) effectively incorporating knowledge of distribution shift into models. With respect to the former, multiple participants proposed using uncertainty quantification to identify and refrain from making predictions for out-of-distribution test examples \citep{hendrycks2016baseline}. Although measurement of distributional divergence was seen as a feasible endeavor, participants noted that interpreting shifts was much more difficult \citep{kulinski2023towards}. With respect to the latter, participants offered an array of modeling methods which were typically motivated by their own applications in the healthcare space. These included propensity-score corrections to adjust for selection bias \citep{abadie2016matching}, as well as more traditional domain-adaptation techniques \citep{gururangan2020don,guan2021domain}. Participants frequently qualified their experiences with a caveat that the methods that worked for them may not work for others. 

Perhaps ironically, a catch-all solution for model generalization did not emerge during the roundtable session. The group was unable to reach a consensus regarding the learning and deployment settings that were most common in practice. Nor were they able to conclusively call for the use of one robust modeling method over another. The only unanimous opinion was that practitioners must thoroughly understand how their training datasets compare to that of their target population. With that knowledge in hand, modeling decisions will inherently vary.
\subsection{Health AI and Accessibility}

\paragraph{Subtopic:} Making AI accessible to all in healthcare is important, but “accessibility” could encompass many things such as infrastructure, compute resources, or access to healthcare in the first place. What are the different components of the healthcare system that could improve patients’ accessibility to health AI, and how do these different components play into the development of AI models?

\paragraph{Chairs:}
\textit{Edward Choi, Kristen Yeom, Edward Lee}

\paragraph{Background:}

Making AI accessible in healthcare encompasses various dimensions from infrastructure and GPU compute resources to train and serve models to the accessibility of models and data to the clinicians and patients. Several components of the healthcare system play an important role in determining the accessibility and effectiveness of health AI. 

\paragraph{Infrastructure and Compute Resources.} Large and robust models require expensive GPUs to train and serve. This also overlaps with data privacy, compliance, and ownership. For example, some institutions will not share patient data over the internet, and therefore developing complex models would necessitate Federated Learning or on-premise training. 

\paragraph{Data Availability and Quality.} AI models are only as good as the data they are trained on. Access to diverse, high-quality, and large-scale datasets is crucial. This includes not only clinical data but also socio-economic, genomic, and demographic data to develop models that can work across heterogeneous datasets. 

\paragraph{Addressing AI accessibility to the patient.} Educating patients about AI in healthcare is crucial for its acceptance and effective use. Patients who understand how AI is used in their care are more likely to engage actively in their health management. 

\paragraph{Ensuring Data Privacy and Security}: Patients are more likely to trust and accept AI-driven healthcare solutions when they are assured their personal health information is secure and used ethically. AI tools must be designed and trained to address the needs of diverse patient groups. This is essential to avoid biases in AI-driven healthcare solutions and to ensure equitable healthcare access.

\paragraph{Discussion:}

Our table focused the core of the discussion on the topic of AI accessibility for researchers and clinicians - specifically the need for larger and more diverse datasets. 

Curating such datasets can be very labor intensive for the clinicians. The process of curating datasets for AI research in healthcare is labor-intensive, especially for clinicians who are already managing significant workloads. Clinicians are not currently incentivized except in an academic setting. Yet, they have to perform careful annotation and categorization. This process is very time-consuming and requires a high level of accuracy and technical expertise to ensure the quality of the dataset.

One participant led the discussion towards the topic of AI accessibility across diverse patient groups. Another participant brought up a paper suggesting that generalizability of a single model on a diverse, heterogeneous population can be overrated. This is primarily because different populations may exhibit distinct medical characteristics, environmental factors, and genetic backgrounds, which can significantly affect model performance. Focusing on personalized models tuned to specific cohorts can be more effective in certain scenarios. These models take into account the unique characteristics and needs of a particular group, potentially leading to more accurate and relevant outcomes for that cohort. Special attention must be given to ensure that cohort-specific modeling does not inadvertently lead to healthcare disparities. Equal effort and resources should be dedicated to developing models for all patient groups, including those that are historically underserved.

\subsection{Health AI and Patient Privacy}

\paragraph{Subtopic:} How can we preserve patient privacy and maintain data security while leveraging machine learning techniques in healthcare?

\paragraph{Chairs:}
\textit{Gamze Gürsoy, Milos Vukadinovic}

\paragraph{Background:} 
To successfully deploy machine learning models in medical practice, we need to train them on diverse and representable large-scale datasets. However, sharing the data between the institutions and countries is bottlenecked by the risk of de-anonymization and harming patient's privacy. Disrespecting the data sharing agreement between the medical provider and patient can have serious implications on the patient's life, and thus cannot be overlooked. While some of these concerns can be partially addressed by ethical and legal measures, in this roundtable we focus on implementing technical measures for preserving patient privacy. Specifically, we focus on securely storing the patient's data, safely deploying medical machine learning models, and minimizing the risk of re-identification.

\paragraph{Discussion:}
We began by exploring real-world cases that highlight the significance of protecting patient privacy in data collected for diagnosis, research purposes, or by direct-to-consumer companies. The examples included The Golden State Killer case \citep{noauthor_golden_nodate}, Latanya Sweeney's identification of the governor of Massachusets health records \citep{noauthor_latanya_2023}, and various linkage attacks.

Building on these examples, we discussed the concept of quasi-identifiers, which are the attributes that when combined with the other data
can identify the patient. Each attribute that carries some information about the patient can be a quasi-identifier, and that is why removing all of them is impossible, instead, we should remove enough so that the risk of re-identification becomes extremely small.
Another way to make re-identification harder is to use differential privacy methods. Differential privacy adds noise to the data to protect the privacy of the patients but aims to preserve the underlying data distribution. While the roundtable recognized the usefulness of differential privacy the most notable insights were that there is no mathematical guarantee of its' privacy and that adding noise in such a way could remove the distribution tails which affects the downstream model accuracy.

In the cases when the medical center does not want to release a dataset or dataset de-identification is impossible, we can train machine learning models using federated learning. The senior chair explained that federated learning is a method where multiple devices or servers train a model locally on their data and then send only the model updates to a central server for aggregation, preserving data privacy and reducing transmission costs. For example, Google successfully deployed federated learning in practice to train their models on the data from users' phones, without sending the raw data to the Google servers. Yet another approach to keep the dataset private is to generate a fully synthetic medical dataset, but this approach comes with its own challenges such as lack of realism, bias, and mode collapse.

Next, the following question was brought up "How can we prevent a machine learning model's weights from revealing its training dataset?".
These attacks are known as model inversion \citep{wang_variational_2022}, and while the risk exists, the roundtable agreed that it takes a lot of sophistication to recover
the original training dataset from the weights only. Nonetheless, the participant noted that we can employ knowledge distillation privacy-preserving techniques such as the teacher-student model to make it even harder for the attacker to retrieve the original data. Namely, the initial model acts as a teacher and the student model learns from the outputs of the teacher model \citep{alkhulaifi_knowledge_2021}, rather than directly from the sensitive training data. 

Finally, we discussed privacy concerns with medical imaging and genetics data. It was mentioned how medical imaging data is easier to de-identify compared to genetics data because usually, the only obvious identifiers are the metadata and text embedded in the image. Genetic data carries a vast amount of information and it is almost impossible to precisely quantify the amount of genetic information leakage in a dataset \citep{gursoy2022genome}. It takes only 50 SNPs (out of $\approx$ 4 million) to uniquely identify the individual \citep{yousefi_snp_2018}, and vast information obtained from genetic data can easily be abused, for example, for calculating polygenic risk score for embryo selection.


\subsection{Bias/Fairness in Health AI}

\paragraph{Subtopic:} Despite its potential, the application of machine learning in healthcare has often resulted in models that reflect and reinforce existing health disparities. How can machine learning promote fairness and enhance global health outcomes?

\paragraph{Chairs:}
\textit{Marzyeh Ghassemi, Emma Pierson, Aparna Balagopalan, Sarah Jabbour}

\paragraph{Background:}
Past research has shown that AI for health may be biased \citep{obermeyer2019dissecting}, and these biases may be introduced into the machine learning pipeline at various stages \citep{chen2021ethical}. During problem selection, data collection, and labeling \citep{spector2021respecting,ferryman2023considering}, existing biases in the healthcare system can influence the generated research datasets, and thus who benefits the most from AI research advancements. During development \citep{gichoya2022ai,pierson2021algorithmic}, AI models might mimic and even exacerbate human biases, resulting in unfair AI predictions. During deployment \citep{verma2023expanding,adam2022mitigating}, AI has been shown to affect patient subpopulations differently as a result of such unfair prediction \citep{paulus2020predictably}, and practitioners face decisions about who receives access to AI \citep{dai2021artificial} (e.g., developing AI that can be widely implemented, versus institution-specific models). 

In this research roundtable, we aimed to discuss how AI researchers can understand the context of, and address AI biases at, these different stages, as well as discuss potential ways forward for fair and responsible deployment. 

\paragraph{Discussion:}
The discussion delved into the multifaceted concept of fairness in the healthcare context, examining aspects such as data collection \citep{spector2021respecting}, parity in outcomes for subgroups \citep{seyyed2021underdiagnosis}, and the entire machine learning (ML) pipeline including deployments\citep{wiens2019no}. Key questions were raised concerning the current state of tools \citep{rajpurkar2022ai}, drawing parallels with the aviation field \citep{bondi2023taking}. The importance of precisely defining the outcome definition \citep{pierson2021algorithmic} and distinguishing between factual and normative questions was highlighted \citep{balagopalan2023judging}. Additionally, the discussion underscored the nuanced differences between bias and fairness \citep{ferrara2023fairness}.

\paragraph{Fairness in the Healthcare Context} A critical point of discussion was the need to audit existing tools developed in the healthcare setting, emphasizing a comprehensive examination of the current situation. The discussion advocated for prospective data collection with a fairness lens. For example, data collection infrastructure for patients outside the hospital was proposed as a means to monitor wound healing, emphasizing the potential impact on healthcare outcomes.

\paragraph{Global Health and AI Deployment} The roundtable conversation also extended to the deployment of tools in settings with limited human experts, citing examples like \acp{LLM} for mental health \citep{cabrera2023ethical}. The discourse questioned whether deploying AI in underserved communities genuinely improves healthcare access. It emphasized the need to consider fairness in terms of access to both models and human experts, cautioning against using \acp{LLM} as a surrogate for human doctors, but rather complementing them ~\citep{abdulhamid2023can}. Machine learning was also presented as a potential tool to expose and illuminate problems for intervention \citep{ferryman2023considering}. The group deliberated about how one might consider using such tools for better interaction in clinical settings, such as aiding in translation between clinicians and patients, or between clinicians and clinicians, who speak different languages. The importance of comprehensive, human-centered evaluations of existing AI technology when adapting them to the clinical setting (e.g., use of translation services in doctor-patient conversations~\citep{mehandru2022reliable}) was emphasized.  

\paragraph{Integration of AI into Healthcare:}
Addressing the fair integration of AI into healthcare, the discussion explored scenarios such as improving documentation over-diagnoses. In the problem selection stage, the importance of strategic thinking from an early stage about the outcomes to be predicted from the lens of health equity was emphasized. Additionally, the impact of (re)thinking the intended deployment settings -- for example, designing AI for a diagnosis prediction task vs aiding in drafting insurance claims -- and continued discussions with clinicians and healthcare workers were highlighted. The challenge of addressing shifts in data due to changing populations \citep{zech2018variable} across hospitals and the complexities of model retraining \citep{otlecs2023updating}, validation, and approval were thoroughly discussed. The healthcare system's lack of careful consideration for staying up-to-date with evolving models and methods, coupled with the absence of FDA approval for clinical risk scores, was highlighted~\citep{harvey2020fda} through the salient example of radiographic severity scores~\citep{pierson2021algorithmic,vina2018natural}. The discussion concluded by acknowledging the inherent challenges in how best to present potentially flawed AI advice~\citep{ghassemi2023presentation}, and emphasized the importance of governance and trust in healthcare systems. Further, the importance of including stakeholders whose workflow would be impacted by the integration of AI -- for example, clinicians and nurses -- in deployment decisions was underscored.

\paragraph{Debiasing in Healthcare:}
The conversation touched upon the intricate relationship between reducing bias and maintaining accuracy \citep{suriyakumar2023personalization, pierson2024accuracy}. Debiasing strategies, such as modeling the truthful distribution of underlying conditions like knee pain, were explored \citep{pierson2021algorithmic}. Relevant examples from translation services, speech recognition, and high-quality synchronous translation services were cited~\citep{mehandru2022reliable}, highlighting the high stakes involved in validating and deploying these technologies~\citep{mehandru2023physician}. The discussion affirmed the significance of improving patient experience in healthcare.
\subsection{ML for Survival Analysis \& Epidemiology/Population Health}

\paragraph{Subtopic:} How can we improve the adoption of novel ML survival models in medical research?

\paragraph{Chairs:}
\textit{George Chen, Sanjat Kanjilal, Vincent Jeanselme}

\paragraph{Background:}
In public health, survival analysis involves modeling the time to an event of interest, offering insights into the effectiveness of interventions, the success of treatments, the identification of vulnerable populations, and the estimation of risk for different conditions. Unlike traditional regression, survival analysis accounts for patients not experiencing the event of interest over the study period. These patients are informed of the probability of observing the event as they are event-free until leaving the study. The outcomes of survival analysis are instrumental in shaping informed policies, guiding treatment strategies, and informing the development of medical guidelines in public health initiatives.

\paragraph{Discussion:}
Despite the development of novel machine learning models for survival outcomes, their application in medical research remains limited, with the field often leaning towards simpler approaches or overlooking the challenge of censoring. This session underscored three critical limitations associated with these novel methodologies, offering directions for future research: (i) interpretability, (ii) data hunger and (iii) disregarding the complexity of real-world medical data.

\paragraph{Interpretability} A critical aspect for medical practitioners is understanding the link between risk factors and outcomes, as this knowledge is essential for informed medical decision-making. Further, in a literature often focused on significance tests, research to quantify credible intervals and uncertainty around the survival predictions is critical for adoption. This exploration can provide a more nuanced understanding of model predictions, enabling more precise targeting of interventions based on the level of uncertainty.

\paragraph{Data Hunger} The reliance on \acp{RCT} and the study of rare conditions frequently leads to limited data availability, a challenge for more flexible models that may struggle under such constraints. Our discussion highlighted the need for research at the intersection of federated learning and survival analysis. This interdisciplinary approach could offer innovative solutions for leveraging distributed datasets without compromising privacy. Additionally, exploring novel methodologies to effectively integrate \acp{RCT} with observational data could open new avenues for overcoming data limitations and enhancing the robustness of survival models.

\paragraph{Medical Complexity} The complexities inherent in medical data, such as unobserved confounders and non-randomized treatment allocation, often pose challenges that novel strategies fail to address. Establishing clear guidelines that articulate the specific challenges each method addresses, along with their associated limitations, is crucial for enhancing the relevance of these models in the medical context. This transparency ensures that medical practitioners are well-informed about the circumstances in which each model is most appropriate, fostering trust and facilitating the integration of novel methodologies into real-world medical practice.

Addressing these limitations will enhance the applicability of survival models to better serve medical research.

\subsection{Causality}

\paragraph{Subtopic:} How can recent advances in AI/ML help discover causal relations using clinical data? To what extent can we use observational data to emulate randomized trials, to evaluate the causal effect of any treatment?

\paragraph{Chairs:}
\textit{Michael Oberst, Linying Zhang, Katherine Matton, Ilker Demirel}

\paragraph{Background:}

Causal inference is prevalent in healthcare; however, it has profound challenges, such as “unobserved counterfactuals,” forcing us to make assumptions about the data-generating process. These assumptions are defensible for the experimental data we collect from the trials, but trials are often costly and not representative of the population. Observational data, such as insurance claims, is large-scale, longitudinal, and more representative. However, causal inference assumptions become dubious with observational data and require expert knowledge to satisfy. We first discuss how we can use observational data to make causal inferences, for instance, by emulating trials, and how the recent advances in machine learning can help leverage high-dimensional and unstructured data such as clinical notes and images.

We then discuss how ideas from causality can improve machine learning algorithms. Traditional machine learning methods learn associations between variables rather than causal relationships. As a result, they are prone to relying on spurious features (aka “shortcuts”) that maximize training performance but do not reliably generalize to changing contexts (e.g., deploying a model at a new hospital). Embedding machine learning models with causal knowledge can help mitigate this problem and produce models that generalize better under distribution shifts. Finally, we explore applications to fairness; for example, how causality can be used to reason about the causal effect of sensitive patient attributes on model decisions.

\paragraph{Discussion:}
We began by discussing challenges related to applying causal inference to real-world clinical data. One challenge is utilizing high-dimensional, unstructured data, such as text and image data. This is difficult because using this data can lead to violations of the “positivity condition”, a key assumption for making valid causal inferences from observational data. The positivity condition requires that for every possible setting of covariates (e.g., patient characteristics) there must be a non-zero probability of observing each treatment condition. When we work with data in very high dimensions, it increases the chance that we observe unique settings of covariates, which violates this assumption. One path to addressing this is to leverage modern machine learning methods for learning low-dimensional feature representations that capture the meaningful aspects of data. Similarly, such methods can help to leverage multi-modal data for causal inference.

Another key challenge in performing causal inference from observational data is evaluation. One solution that was brought up is to use synthetic data to perform sensitivity analyses. However, it is challenging to create realistic synthetic data, so the conclusions drawn from this approach may be limited. An alternative is to use “knowledge-based” sensitivity analysis. One strategy in this realm is to include “negative control outcomes,” or outcomes that are known in advance to not be causally affected by the treatment. This can serve as a sanity check —  if the treatment appears to affect the negative control outcome, this suggests that our causal inference estimates are not valid. Finally, we discussed strategies for evaluation when we have both observational and experimental data. In this case, it may be of interest to use the observational data to estimate treatment effects for sub-groups that are not present in the RCT data. To assess the validity of these estimates, we can test how well the assumptions we make hold for the groups that are represented in both the experimental and RCT data.

We then moved to discuss how ideas from causality can be used to improve machine learning. There has been a lot of excitement in the field lately about the idea of embedding machine learning models with casual reasoning as a way of improving their robustness. We started by interrogating what this means — one participant asked the question “What does it mean for a machine learning model to have casual reasoning?” We discussed how this idea is often too general and vague to be tractable. For example, when classifying diseases from chest X-rays, what would a machine learning model with causal reasoning look like? It is hard to say given that it’s difficult to reason about the causal effect of a disease on images in the pixel space. We discussed how a more reasonable approach for ``causally-informed ML" may be to try to enumerate a set of possible changes that we want a machine learning model to be invariant against. We can then use causality to reason about how well the model will perform under these changes.

Finally, we discussed the implications of causality for fairness, and in particular, how causality can help to assess the fairness of clinical treatment decisions. For example, we can employ counterfactual reasoning to consider how a treatment decision would have been different if some aspect of the patient (e.g., race, gender, etc.) was different. This can help reveal cases of biased and unfair clinical decision-making.

\section{Summary}

The themes from the eleven research roundtables were extracted and grouped into four general areas: (1) foundation models, (2) incorporating domain expertise in model building, (3) addressing fair and robust ML for health concerns, and (4) considering safety, data sharing, and hospital infrastructure throughout the ML pipeline. Here, we summarize these discussions in an effort to highlight recent advancements in machine learning for healthcare for researchers in this domain. 

The discourse surrounding foundational models in healthcare underscores the profound opportunity to utilize large-scale machine learning models for mining the extensive, yet largely untapped, datasets generated within healthcare environments. Despite the hurdles in data consolidation across medical facilities, attributed to disparate standards and privacy concerns, the forward march of AI technology, especially with the advent of \acp{LLM}, offers promising avenues for improving healthcare through the analysis of diverse data types, including clinical notes, images, and genetic information. Furthermore, many recent works adopt approaches to address the multimodal nature of medical datasets including text, images, audio, video, and omics data, opening the door to refined diagnostic accuracy and tailored treatment strategies. Nonetheless, this approach also presents challenges in data stewardship, ethical governance, and the imperative for sophisticated models capable of deciphering this multifaceted data landscape. 

Leveraging domain expertise for Health AI encapsulates a nuanced understanding of integrating AI into healthcare, emphasizing the importance of clinician-AI collaboration, effective integration into clinical workflows, and the development of regulatory frameworks that foster innovation while ensuring safety. Challenges in data acquisition, model evaluation, and the necessity for AI tools that are to be adopted continuously to dynamic healthcare environment were highlighted, alongside the potential of novel machine learning models in survival analysis and epidemiology to refine medical research. However, it was noted that there still exist hurdles related to interpretability, data scarcity, and the complexity of medical data. One way to address these challenges is exploring causality in AI/ML to discern causal relationships from clinical data and emulate randomized trials through observational data. This holds promise for enhancing model robustness, and fairness in clinical decision-making, and ensuring equitable treatment. These discussions illuminate the critical role of domain expertise in advancing Health AI, underscoring the need for technological innovation, regulatory foresight, and collaborative synergy among all stakeholders to unlock the transformative potential of AI in healthcare.

Addressing questions around bias and fairness was discussed at several roundtables. Bias can be introduced into AI at various stages of the development pipeline, such as data collection and labeling. Addressing such bias during model development could come in many different forms, such as precisely defining the outcome of interest and prospectively collecting data with a fairness lens. Fairly integrating AI into healthcare also requires careful consideration, such as increasingly staying up to date with evolving models and methods and improved governance. Another component of making AI fair is making it accessible to all patients, which encompasses many different parts of the AI pipeline: hospital infrastructure, compute resources, and data availability. Ensuring the ethical development of AI, including preserving patient data and privacy, can be a way to increase AI adoption by both patients and healthcare providers. Preserving patient privacy can be achieved through both legal measuring and technical measures, which was the focus of another roundtable. For example, techniques such as removing enough data can ensure that re-identification via quasi-identifiers or differential privacy methods cannot be employed. With recent advancements in generative AI, there is increasing focus on mitigating model inversion (i.e., machine learning model weights revealing training datasets). 

Addressing dataset shift and model generalizability across hospitals and patient populations was also discussed by the roundtables. One discussion focused on how diverse medical settings, differences and changes in hospital protocols, and heterogeneous patient populations contribute to the challenging endeavor of integrating AI across institutions. Another table discussed how effectively adapting to such shifts requires addressing several questions around characterizing distribution shifts and building models resilient to them, though healthcare poses unique challenges that require novel solutions to such questions.

\section{Lessons Learned}

Throughout the preparation of ML4H 2023 Research Roundtables, we followed topic organizations and instructions on recruiting chairs from the previous year's roundtable document \citep{hegselmannrecent}. Recruiting junior and senior chairs posed a significant challenge this year, and many junior chairs were added after paper decisions, leading to many last-minute inclusions. One of the challenges includes an organizational shift from hybrid conferences to the first full in-person conference. The number of participants for different topics varied significantly, but on the first round of roundtable discussion participants were distributed more evenly throughout the topics. Post-conference, each chair provided a summary paragraph, which helped in the creation of this report. The summary document from last year helped form guidelines for topic summaries. 

Overall, roundtable sessions were successful with active involvement, resembling a panel discussion but with a smaller scale and informal setting that helped networking among attendees and experts on the topic. We recommend ML4H 2024 conference to incorporate roundtable sessions to boost engagement and networking prospects.

\bibliography{references}

\end{document}